# A Fuzzy-Enhanced Explainable AI Framework for Flight Continuous Descent Operations Classification

Amin Noroozi, Sandaruwan K. Sethunge, Elham Norouzi, Phat T. Phan, Kavinda U. Waduge, and Md. Arafatur Rahman, *Senior Member, IEEE*

*Abstract*— Continuous Descent Operations (CDO) involve smooth, idle-thrust descents that avoid level-offs, reducing fuel burn, emissions, and noise while improving efficiency and passenger comfort. Despite its operational and environmental benefits, limited research has systematically examined the factors influencing CDO performance. Moreover, many existing methods in related areas, such as trajectory optimization, lack the transparency required in aviation, where explainability is critical for safety and stakeholder trust. This study addresses these gaps by proposing a Fuzzy-Enhanced Explainable AI (FEXAI) framework that integrates fuzzy logic with machine learning and SHapley Additive exPlanations (SHAP) analysis. For this purpose, a comprehensive dataset of 29 features, including 11 operational and 18 weather-related features, was collected from 1,094 flights using Automatic Dependent Surveillance-Broadcast (ADS-B) data. Machine learning models and SHAP were then applied to classify flights' CDO adherence levels and rank features by importance. The three most influential features, as identified by SHAP scores, were then used to construct a fuzzy rule-based classifier, enabling the extraction of interpretable fuzzy rules. All models achieved classification accuracies above 90%, with FEXAI providing meaningful, human-readable rules for operational users. Results indicated that the average descent rate within the arrival route, the number of descent segments, and the average change in directional heading during descent were the strongest predictors of CDO performance. The FEXAI method proposed in this study presents a novel pathway for operational decision support and could be integrated into aviation tools to enable real-time advisories that maintain CDO adherence under varying operational conditions.

## I. INTRODUCTION

Continuous Decent Operations (CDOs) are advanced arrival procedures where an aircraft descends continuously from cruise altitude to final approach with minimal engine thrust and few level-offs, thereby reducing fuel burn, $CO_2$ emissions, and noise [1]. Cruise altitude refers to the specific and usually constant height above sea level at which an aircraft maintains steady flight for the majority of its journey after climbing from takeoff and before starting its descent. Level-off also refers to the period during descent when an aircraft temporarily stops descending and flies horizontally at a fixed altitude. The arrival route, defined by a Standard Terminal Arrival Route (STAR), guides an aircraft during the descent phase from cruise altitude toward the airport. This route consists of a sequence of waypoints and altitude constraints designed to ensure safe and efficient traffic flow. The overall arrival trajectory is divided into multiple flight segments, each corresponding to a portion of the route the aircraft follows between two consecutive waypoints. Guidelines published by the International Civil Aviation Organization (ICAO) have formalized CDO definition, classifying a descent as a true CDO if it contains no or minimal intermediate level-off segments [1], [2]. In practice, however, many arrivals deviate from the ideal CDO profile due to operational constraints. While considerable progress has been made in designing optimized STARs and descent profiles, operational variations frequently hinder achieving the full benefits of CDO procedures. Such variations include uncertainties such as unpredictable pilot behavior, real-time Air Traffic Control (ATC) instructions, and varying weather conditions, all of which can lead to deviations from the ideal descent profile [3], [4].

Achieving optimal CDO profiles is a key goal in modern air traffic management due to the significant environmental and operational benefits, including reducing fuel consumption, emissions, and noise pollution during the aircraft descent phase [1]. Continuous descent, significantly reduces the operational inefficiencies traditionally associated with stepped descent approaches [5]. Consequently, effective evaluation and prediction of flights' CDO adherence, defined as the proportion of CDO-compliant flight segments to the total number of segments within the arrival route, remains crucial for optimizing aviation operations, enhancing efficiency, and supporting environmental sustainability. Accurate and interpretable CDO adherence classification enables airspace planners and operators to systematically assess descent performance, prioritize operational improvements, and develop targeted interventions aimed at enhancing sustainability and operational efficiency.

Recent advancements in Machine Learning (ML) techniques have facilitated the optimization of aviation operations, particularly by leveraging trajectory data extracted

(Corresponding author: Amin Noroozi).

Amin Noroozi, Phat T. Phan, Kavinda U. Waduge, and Md. Arafatur Rahman are with the School of Engineering, Computing and Mathematical Sciences, University of Wolverhampton, Wolverhampton WV1 1LY, United Kingdom (e-mail: a.noroozifakhabi@wlv.ac.uk)

Sandaruwan K. Sethunge was with the School of Engineering, Computing and Mathematical Sciences, University of Wolverhampton, Wolverhampton WV1 1LY, United Kingdom. He is now with Qatar Civil Aviation Authority, Doha 7GQW+6QP, Qatar (e-mail: pubudusanda@gmail.com)

Elham Norouz was with the University of Surrey, Guildford GU2 7XH, United Kingdom. She is now an independent researcher, London, United Kingdom (e-mail: elhamnorouzi85@gmail.com),

from Automatic Dependent Surveillance–Broadcast (ADS-B) systems [6], [7], [8], [9]. ADS-B is a surveillance technology used in air traffic management that enables aircraft to broadcast their positional and performance information, such as latitude, longitude, altitude, velocity, and heading, in real-time to ground stations and other aircraft. This broadcast occurs at regular intervals providing continuous and precise trajectory data that can be utilized for monitoring, analyzing, and optimizing flight operations [6].

While ML models have been used for applications such as flight trajectory prediction and optimization [7], [8], [10], flight scheduling [11] and Air Traffic Management (ATM) [12] identifying flight phases [6], there is currently a lack of studies classifying flights' full CDO adherence. Although trajectory optimization provides a framework for improving aircraft performance across all flight phases, optimizing CDO adherence often requires a dedicated approach. This is because the descent phase presents unique operational constraints and goals, such as minimizing noise and emissions near airports, adhering to complex airspace restrictions, and accommodating real-time air traffic variations, which are not adequately addressed when using a generalized, full-flight trajectory optimization model [13], [14]. CDO optimization allows for tailored consideration of these descent-specific factors, leading to more practical, environmentally effective, and operationally robust solutions for approach and landing procedures. Furthermore, optimizing CDO separately allows for more targeted interventions, such as fine-tuning descent profiles and integrating with ATC arrival management systems, which are often not captured in holistic trajectory optimization models [15].

Existing ML methods used for the applications mentioned above still suffer from limited interpretability, significantly restricting their acceptance in safety-critical domains like aviation, where transparency and explainability are essential for stakeholder trust and operational adoption. Previous studies have highlighted the importance of XAI in enhancing human factors such as situation awareness, trust, and decision-making when working with AI tools in aviation optimization [16], [17]. Nevertheless, ensuring operational interpretability, that is, explanations understandable to aviation experts and consistent with known guidelines, remains a challenge for adoption of ML models in aviation systems.

To address this challenge, this study proposes a hybrid Explainable Artificial Intelligence (XAI) approach to classify flights' adherence to CDO procedures using operational ADS-B trajectory and weather-related data. For this purpose, we collected a comprehensive dataset comprising 1094 flights descending into Doha Terminal Maneuvering Area (TMA). From this dataset, we then engineered a robust set of 11 trajectory-based operational features directly related to CDO performance, which represent altitude changes, speed profiles, direction variations, number of trajectory segments, and the distance traveled within the TMA. We additionally included 18 features related to the weather condition to account for this uncertainty as highlighted by previous studies [4]. The 29 gathered features were finally used to classify flights' CDO adherence level. Subsequently, we applied SHapley Additive exPlanations (SHAP) analysis [18], [19] to identify the most influential features driving CDO classification.

While SHAP provides valuable insights into feature importance, its effectiveness in explaining the behavior of models trained on structured, tabular data can be limited, as numerical attributions often lack a human-readable logical structure. Although SHAP values quantify individual feature contributions, they do not inherently represent if-then rules or specify the conditions under which predictions change, elements often essential for operational transparency and interpretability. Consequently, several recent studies have argued that such methods do not necessarily lead to a deeper understanding of a model's internal reasoning [20], [21]. To address this limitation and enhance the model's explainability, we employed the top three features identified by SHAP to construct a fuzzy inference system and derive a set of interpretable fuzzy rules [22], [23], [24]. These rules, expressed as human-readable if-then statements, encapsulate the model's decision-making logic in a transparent and intuitive manner, providing a clear and structured explanation that is easily understood by domain experts and operational users. The proposed method is referred to as the fuzzy-enhanced explainable artificial intelligence (FEXAI) throughout the remainder of the paper. Fig. 1 illustrates the different CDO levels and the objective of FEXAI.

FEXAI was rigorously validated against the collected flight data, demonstrating not only reliable and accurate predictive performance but also explicit reasoning behind CDO classification decisions. The transparency provided by this approach enables more informed discussions between air traffic controllers, pilots, airline operations centers, and regulatory bodies, thereby enhancing collaborative efforts toward optimized descent procedures and improved environmental outcomes.

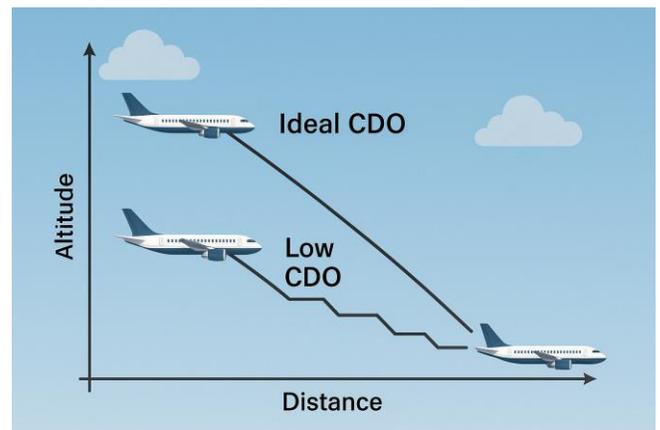

**Fig. 1**. A flight with an ideal CDO, showing no intermediate level-off segments, compared to a flight with a low CDO that includes multiple level-off segments. The goal of FEXAI is to predict flights' CDO performance using operational and weather features and to explain the reasoning behind the predictions through human-readable If–Then rules.

Based on the discussion above, the contributions of this paper are fourfold:

1- The first key contribution of this study is the collection of an extensive ADS-B-based dataset for CDO evaluation. To the best of our knowledge, this is the first comprehensive dataset designed for this purpose, comprising 29 diverse features, including both operational and weather-related variables.
2- The second contribution is defining and classifying the CDO adherence using the collected data. As discussed above, there is currently a lack of methods that directly classify CDO adherence, despite its importance.
3- The third contribution of this study is the systematic identification of influential trajectory features using SHAP analysis, which offers data-driven insight into the factors most associated with CDO adherence.
4- The fourth major contribution is the development and validation of the FEXAI framework, which generates explainable fuzzy rules from the top-ranked features identified by SHAP. This approach provides an interpretable and operationally meaningful framework for assessing and optimizing continuous descent operations.

The remainder of the paper is structured as follows: Section II reviews previous studies related to CDO optimization and the application of ML in this domain. Section III provides a detailed description of the proposed methodology. The results and discussion are presented in Sections IV and V, respectively. Finally, Section VI concludes the paper.

II. RELATED WORK

While ML methods have been explored in trajectory optimization, arrival scheduling, and ATM, there is currently a lack of studies directly applying ML to CDO optimization. Traditional approaches mainly focus on conventional methods for trajectory optimization and arrival scheduling to enable as many CDOs as possible. This includes techniques like dynamic arrival route design, speed/path stretching, and sequencing algorithms to ensure sufficient spacing between aircraft for continuous descents [25], [26], [27]. However, in recent years, the emphasis has shifted toward data-driven and ML techniques to assist trajectory optimization [7]. ML and Deep Learning (DL) can leverage large amounts of flight data to predict or classify descent trajectories and suggest optimized procedures that account for the inherent uncertainty in operations [28]. For example, one study proposed an adaptive model to predict level flight time uncertainty, for ground-based 4D trajectory management, as a function of flight and meteorological conditions, including Mach number, flight distance, wind, and temperature [29]. 4D trajectory refers to a flight path that includes three spatial dimensions (latitude, longitude, altitude) plus time as the fourth dimension. It defines not just where an aircraft should be, but when it should be there. The proposed model in their study demonstrated the capability to dynamically adjust its flight time estimates based on various weather factors, such as wind and temperature, as well as aircraft performance.

Another study introduced a long-term 4D trajectory prediction method using generative adversarial networks (GAN) [10]. For this purpose, the authors designed three predictive models by integrating a GAN with a one-dimensional convolutional neural network (CNN), a two-dimensional CNN, and a long short-term memory (LSTM) network, respectively. They then used historical 4D trajectory data from Beijing to Chengdu, China, to predict the aircraft's long-term trajectory. In the end, they concluded that the one-dimensional CNN performs more accurately than the other two models. Investigators [30] proposed a two-step ML model to improve the selection of the Top-of-Descent (TOD) point, which is the point where the aircraft begins its descent toward the destination airport. In their approach, a Random Forest classifier was trained on historical radar data to learn how various factors (relative positions, altitudes, and convergence of neighboring traffic) influence whether a flight can perform a continuous descent. Results on a large set of real arrival data from Singapore flight information region (FIR), using the ATM data for November 2019 showed that the ML-predicted TOD was within ±10 nautical miles of the actual (controller-chosen) TOD in over 88% of cases.

A few studies have also used ML and DL to estimate the aircraft fuel consumption to achieve optimal descent profiles [31]. Using flight data recorder (FDR) and ADS-B data, authors in [32] developed and evaluated neural networks and Classification and Regression Tree (CART) models to predict the fuel consumption. In the end, they reported an estimation error of 6.3% and 14.1% for CART and neural networks, respectively. Another study [33] proposed a covariance bidirectional extreme learning machine, referred to CovB-ELM, to predict the flight trajectory and resulting fuel consumption. Using real flight historical high-dimensional data from an international airline in Hong Kong, they reported significant improvement in the prediction error compared to the existing conventional methods used in airlines. A Radial Basis Function (RBF) neural networks was proposed in [34] to predict flight fuel consumption. They used high-resolution onboard Quick Access Recorder (QAR) flight data, to develop the RBF model for different flight phases, including takeoff/climb, cruise, and descent/approach. The results demonstrated an error of 5.73%, 3.36%, and 14.04% in predicting takeoff, cruise, and descent phases, respectively, improving the performance compared to the existing models.

While ML and DL methods have been explored in trajectory optimization, XAI remains underutilized in the literature, with its application mostly limited to ATM problems primarily in domains such as conflict detection and resolution, aviation safety, air traffic flow management, and delay prediction [35]. One study [36] built supervised learning models, namely, Support Vector Machine (SVM), Logistic Regression (LR), Random Forest (RF), and Deep Neural

Network (DNN) to classify aviation safety outcomes in Australia into three classes, Accidents, Incidents, and Serious Incidents, and used SHAP to identify the key features influencing the model's decisions. In the end, they found that when their dataset was balanced, all models assigned high importance to intuitively relevant features like latitude and longitude of occurrence and the operational category of flight. A recent study [37] introduced a "self-paced ensemble" model for predicting significant air turbulence events near airports (caused by wind shear), and they employed SHAP to interpret the model's outputs. The SHAP analysis revealed that features related to wind shear, specifically the altitude of the shear and its magnitude, were the most influential factors for the turbulence predictions. Multiple studies have also highlighted XAI approaches as essential for trustworthy, fair, and transparent AI systems in aviation, including ATM [17], [38].

Fuzzy logic has been a popular approach for handling uncertainty and imprecision in control systems and decision-making for transportation and aviation systems [39] and has been applied in various ways, from control of individual aircraft [40] to higher-level air traffic management and trajectory tracking [39]. Fuzzy logic provides a framework to encode human expert knowledge in the form of linguistic if-then rules and truth values that range between 0 and 1, rather than the binary true/false of conventional logic. The fuzzy system helped design and adjust the Space Shuttle's descent trajectory in real time, accounting for "unusual energy states" (deviations in altitude/velocity) and even potential control surface failures [41]. One study [40] proposed an optimized sequence control fuzzy logic controller (FLC) for aircraft landing gears and reported a 25% and 15% improvements in the accuracy of transitions and response time, respectively, compared to the conventional systems. Investigators [39] proposed a fuzzy logic algorithm to track aircraft trajectory and reported that the fuzzy tracker outperformed a standard Kalman filter when tracking aircraft with uneven accelerations and sudden direction changes. Fuzzy logic has been applied to conflict detection and resolution in air traffic control. Authors [42] used a "fuzzy reasoning" approach to evaluate the degree of conflict between aircraft trajectories and to suggest conflict-resolving maneuvers. For this purpose, they combined fuzzy logic with genetic algorithms to resolve aircraft conflicts, encoding ATC heuristics, such as preferences for altitude vs. heading changes, as fuzzy rules and then optimizing the solution. Such a system can output advisories like "reduce speed slightly" or "turn a few degrees right", which controllers can then interpret and implement. Despite limited literature directly addressing fuzzy logic in CDO optimization specifically, related aviation studies suggest the potential for fuzzy approaches to improve robustness and adaptability in descent trajectory planning and optimization [40], [41].

As evidenced by the studies reviewed above, despite the significance of CDO and its benefits, datasets for CDO optimization remain scarce in the literature. Consequently, there is a critical shortage of methods designed to optimize CDO and identify the factors influencing its performance. Furthermore, most existing approaches in related applications lack the level of explainability required in aviation. The following sections introduce the new dataset and the FEXAI framework developed to address these gaps.

III. METHODS

Fig. 2 presents an overview of the methodology and the proposed FEXAI framework. The process begins with data collection and the compilation of a new dataset containing operational and weather features. The FEXAI framework is then implemented in five stages:

**Step 1**: Three classifiers, Random Forest (RF), XGBoost (XB), and CatBoost (CB), are applied to predict CDO adherence levels using all features in the dataset; these are referred to as ML classifiers in this paper [43], [44], [45].

**Step 2**: SHAP analysis is conducted to rank the features.

**Step 3**: The top three features with the highest SHAP scores are selected.

**Step 4**: A fuzzy classifier, referred to as the FEXAI classifier, is constructed and applied to classify CDO adherence levels using only the top three features.

**Step 5**: Explainable fuzzy rules are extracted from the outputs of the FEXAI classifier.

*A. Data Collection*

Trajectory data for 1,094 flights were collected Flightradar24[1], a publicly available commercial ADS-B data portal. This platform aggregates ADS-B broadcasts received via ground-based receivers, providing positional, speed, altitude, and heading data for aircraft in real time. The selected flights were all inbound to Doha International Airport, with trajectories corresponding to published STARs. Data was collected for months February to September 2024, covering both peak and off-peak traffic hours. Only flights entering Doha TMA from the North and East sectors were included; arrivals from the South and West sectors were excluded due to an insufficient sample size within the data collection period.

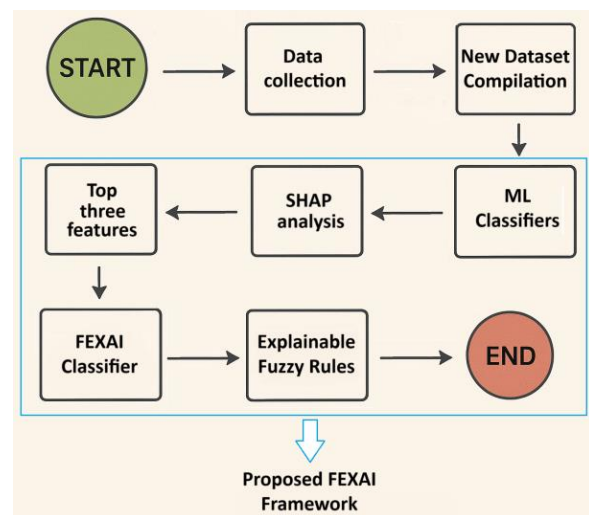

**Fig. 2**. The methodology and proposed FEXAI framework

[1] https://www.flightradar24.com

For this study, only data related to the arrival route, which includes segments within the Doha TMA, were retained by filtering trajectories based on geographic boundaries. The arrival phase of a flight begins at the boundary of the TMA, where designated waypoints mark intersections of air routes. Upon reaching these waypoints, the aircraft follows a predefined arrival route defined by STAR. Accordingly, we started collecting arrival data from these initial waypoints. Additionally, as STAR procedures typically conclude at altitudes between 2,500 ft and 3,500 ft [1], flight data recorded below 3,500 ft were excluded to ensure that the dataset fully represents the arrival phase under STAR procedures. Fig. 3 depicts the arrival route of two flights from Doha TMA boundary to STAR termination waypoint.

*B. New Dataset Compilation*

Using the collected raw ADS-B data, we compiled a new dataset consisting of 11 operational and 18 weather features. The following operational features were extracted from the ADS-B data:

**Sector**: This feature categorizes the entry point of the arrival route into North or East sectors and is determined based on the aircraft's geographic entry point coordinates (Latitude/Longitude) as it enters the Doha TMA.

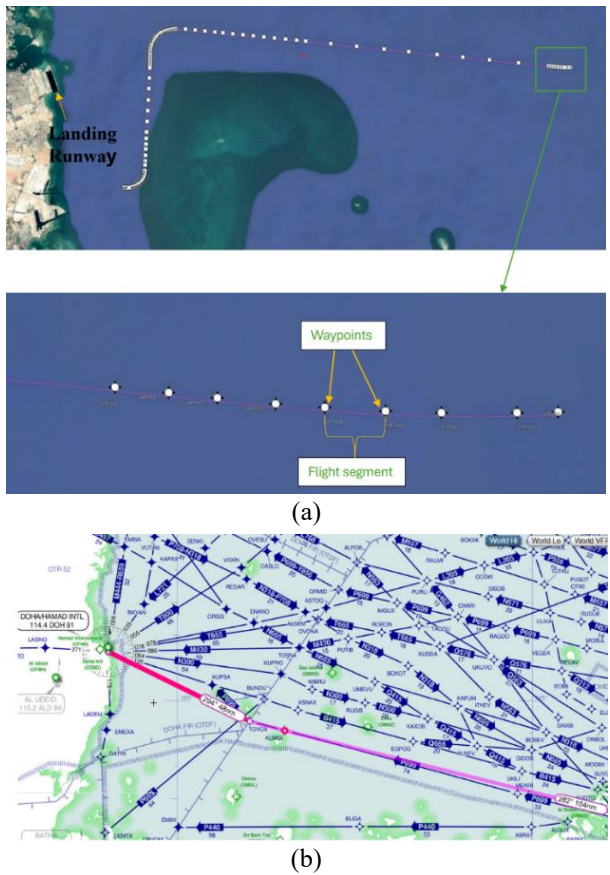

**Fig. 3.** (a) Arrival route of a flight from TMA boundary to STAR termination waypoint and corresponding flight segments. (b) A real flight path to Doha TMA (purple line) and the arrival route (red line).

**Altitude:** The altitude (in feet) of the aircraft at the entry point of the arrival phase within Doha TMA, directly collected from ADS-B data.

**MSpeed:** The average ground speed (i.e. the actual speed relative to the ground) of the aircraft across all segments within the TMA, expressed in nautical miles per hour (knots).

**MDRate:** The average descent rate across all segments within the TMA, expressed in feet per nautical mile (ft/NM).

**FltSegments:** The total number of flight segments within the TMA extracted from ADS-B data.

**Distance_NM:** Total horizontal distance in nautical miles covered within the TMA.

**MDirection:** Average change in heading angle in degrees between consecutive segments. This feature measures how much the aircraft changes its track along the STAR, indicating trajectory complexity.

**StartLati:** Latitude of the first recorded point within the TMA for each flight.

**StartLong:** Longitude of the first recorded point within the TMA for each flight.

**EndLati:** Latitude of the final recorded point within the TMA, before 3,500 ft

**EndLong:** Longitude of the final recorded point within the TMA, before 3,500 ft.

In addition to the operational features, we used the flight timestamped data to retrieve the following nine weather features from a weather API[2] for the entry and exit points of the arrival route (labelled as start and end, respectively), resulting in 18 features in total:

**start_temp** and **end_temp**: the ambient air temperature (in °F).

**start_feels_like** and **end_feels_like**: the "feels-like" temperatures (in °F).

**start_pressure** and **end_pressure**: the atmospheric pressure (in hPa).

**start_humidity** and **end_humidity**: the relative humidity percentage (%).

**start_dew_point** and **end_dew_point**: the dew point temperatures (in °F).

**start_clouds** and **end_clouds**: the cloud coverage percentage (%).

**start_wind_speed** and **end_wind_speed**: the wind speed (in miles per hour, mph)

**start_wind_deg** and **end_wind_deg**: the wind direction (in degrees).

**start_weather** and **end_weather**: descriptive weather conditions (e.g., clear, cloudy, rain).

The arrival route usually covers a distance roughly between 20 and 50 nautical miles [46], providing enough time and distance for a controlled descent from cruising altitude to an altitude near 3500 feet where the instrument approach begins. Since weather conditions may vary along this distance, including weather data for both the entry and exit points of the arrival route ensures that such uncertainty is adequately addressed.

---

[2] https://openweathermap.org/api

CDO adherence level for each flight was calculated as follows:

$$\text{CDO}_{\text{adherence}} = \frac{FltSegments^{CDO}}{FltSegments} \quad (1)$$

where $FltSegments^{CDO}$ and $FltSegments$ show the number of CDO compliant flight segments and the total number of flight segments within the TMA boundary, respectively. A categorical variable, CDOCAT, was then created by encoding $\text{CDO}_{\text{adherence}}$ as follows.

$$CDOCAT = \begin{cases} Low & \text{CDO}_{\text{adherence}} < 30\% \\ Medium & 30\% < \text{CDO}_{\text{adherence}} < 55\% \\ High & \text{CDO}_{\text{adherence}} > 55\% \end{cases} \quad (2)$$

Since there is no universally established standard categorizing CDO adherence levels, these thresholds were empirically defined, considering previous studies that reported average CDO compliance margins of approximately 30% for conventional descents and around 60% for managed or optimized CDO procedures [47]. The encoded CDOCAT variable was finally used as the target variable in the classification stage.

*C. ML Classifiers and SHAP Analysis*

The 29 extracted features, comprising 11 operational features and 18 weather-related features, were used as input to three ML classifiers, including RF, XB, and CB, to predict the target variable, CDOCAT. To explore different classification strategies, we considered three scenarios. In the first scenario, the original three-class formulation of CDOCAT (Low, Medium, High) was retained. In the second scenario, the Medium and High classes were merged into a single class labelled "Not Low," resulting in a binary classification task distinguishing between Low and Not Low. In the third scenario, the Low and Medium classes were merged into a "Not High" class, yielding another binary classification between Not High and High. These scenarios were designed to assess the models' predictive ability under both multi-class and binary settings, with particular focus on use cases where targeted binary classification, such as identifying only Low-risk or High-risk cases, is more practical or informative.

For each classification scenario, four evaluation metrics, accuracy, precision, recall, and F1 score, were computed using five-fold cross-validation. The best-performing model from each scenario was then used to compute SHAP values for feature importance analysis. The SHAP analysis was conducted in two stages:

1. **Global Feature Importance Estimation**: In the three-class scenario, SHAP values were computed separately for each class, with their absolute values averaged across classes, test samples, and folds, and were then normalized to obtain cross-validated global importance scores. For the binary classification scenarios, absolute SHAP values were averaged directly across test samples within each fold, then across folds, and normalized to produce the global importance scores.

2. **Class-Specific SHAP Analysis**: In the second stage, a class-specific SHAP analysis was conducted using five-fold cross-validation. This analysis focused on the top three features with the highest global importance scores from the first stage, in order to examine the direction and magnitude of their contribution toward the classification.

To assess feature-level class separability in the two binary classification scenarios (i.e., the second and third scenarios), we additionally computed the Wasserstein Distance (WD), also known as the Earth Mover's Distance [48], between the SHAP value distributions of each feature across the two classes. WD is a non-negative real-valued metric that quantifies the distance between two probability distributions by calculating the minimum "cost" required to transform one distribution into the other. In this context, higher WD values within each scenario indicate greater divergence between the class-specific SHAP value distributions, reflecting stronger separability in how the model differentiates between classes, thereby a stronger classification.

*E. FEXAI classifier and Explainable Fuzzy Rules*

The top three features with the highest SHAP importance scores were used as inputs to a simplified fuzzy rule-based winner-take-all classification system [49], referred to as the FEXAI classifier. This system was applied to classify the CDOCAT variable and generate interpretable fuzzy If–Then rules as follows:

1- **Fuzzification**: Each numerical input feature (e.g., MDRate) was mapped to expert-defined fuzzy sets (e.g. Low, Medium, High). For each input value, the fuzzy set with the highest membership degree was selected (winner-take-all selection) to simplify rule extraction and improve interpretability.

2- **Rule Extraction**: For each training sample, a single fuzzy IF–THEN rule was generated by combining the selected fuzzy sets for the three features. For example, "IF MDRate is High AND FltSegments is Few AND MDirection is Complex THEN CDOCAT is Not Low" was generated when such conditions occurred in the training data.

3- **Inference and Prediction:** The extracted fuzzy rules formed the rule base to infer (predict) CDOCAT. For each test sample, the system matched its input to the corresponding rule antecedent and returned the associated fuzzy output.

4- **Defuzzification**: The fuzzy system output from the previous step (in the range [0, 1]) was thresholded (e.g., at 0.5) to obtain a crisp binary class label.

5- **Evaluation**: The predicted labels were compared with the ground truth to compute evaluation metrics, assessing the predictive performance of the extracted rule set.

The FEXAI classifier and rule extraction were applied only to the best-performing classification scenario from the previous step, which, according to the results, was the binary classification distinguishing Low from Not Low CDO adherence. Fuzzy rules and evaluation metrics were derived using a five-fold cross-validation procedure. In each fold, rules were generated based on the top three features ranked by SHAP importance, and the final rule set was obtained by taking the union of all unique rules generated across folds.

IV. RESULTS

*A. ML Classification*

In this section, we present the results of the first step in the FEXAI framework, namely ML classification. In this step, all features in the dataset were used to classify CDO adherence with the ML classifiers, RF, CB, and XB. The FEXAI classifier is not applied at this stage because the features have not yet been ranked.

Table I presents the results for all classification scenarios. In each scenario, the bold values indicate the best performance for each evaluation metric among the classifiers. Overall, all classifiers achieved high accuracy, with the CB classifier outperforming others in most cases. Notably, the second classification scenario (Low vs. Not Low) yielded the most consistent performance across all classification scenarios, with all performance metrics for all classifiers exceeding 90% and, in most cases, surpassing 91%.

Since the CB classifier demonstrated the best performance, and for the sake of simplicity, we used only this classifier for the SHAP analysis.

*B. SHAP Analysis and Top Three Features*

This section presents the results of the SHAP analysis using the CB classifier. Fig. 4 shows the global feature importance in different scenarios using the CB classifier. MDRate, FltSegments, and MDirection are consistently ranked among the top three features, with significantly higher SHAP scores compared to other features.

Fig. 5 presents the class-specific SHAP values for these top three features in the second scenario, which achieved the best classification performance, Low vs Not Low, with Not Low treated as the positive class. As shown, MDRate and MDirection values approximately greater than 0.02 and 1, respectively, tend to push predictions toward Not Low CDO adherence, whereas FltSegments values below approximately 200 have the same effect. This indicates that flights with higher mean descent rate (MDRate > 0.02), greater directional changes during arrival (MDirection > 1), and fewer arrival-phase segments (FltSegments < 200) are more likely to exhibit better CDO performance.

For MDRate between ~0.02 and 0.035, SHAP values increase linearly, meaning that higher descent rates within this range strongly improve CDO adherence likelihood. Between ~0.035 and 0.06, SHAP values plateau, suggesting diminishing returns in CDO improvement. Beyond ~0.06, SHAP values decrease slightly (while remaining positive), implying that very steep descent rates, which may occur under atypical operational conditions (e.g., late clearances, weather avoidance), can reduce CDO efficiency.

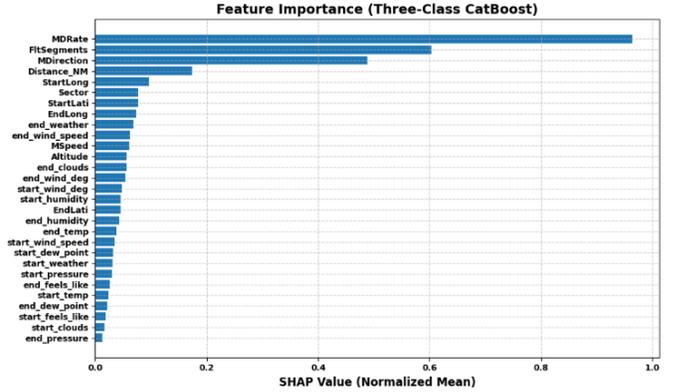

(a)

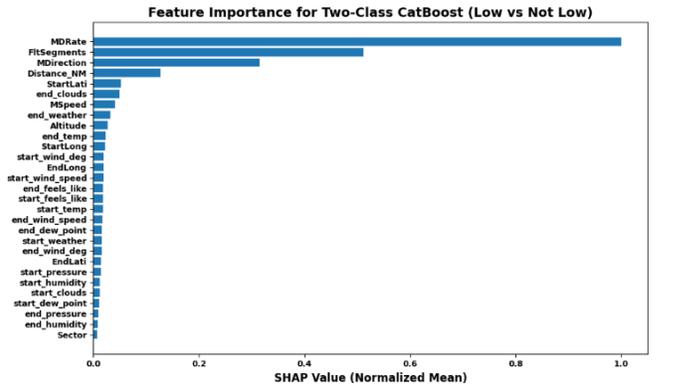

(b)

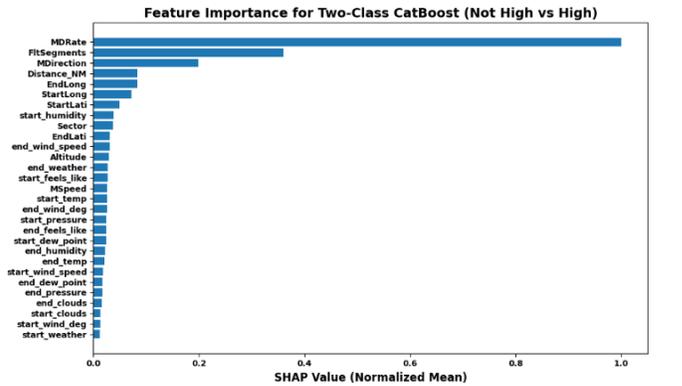

(c)

**Fig. 4**. The global SHAP feature importance for (a) three-class, (b) two-class (Low vs Not Low), and (c) two-class (Not High vs High) classifications.

TABLE I
CLASSIFICATION PERFORMANCE (%) USING DIFFERENT SCENARIOS AND ML CLASSIFIERS

| Scenario | Classifier | Acc | Pr | Recall | F1 score |
|---|---|---|---|---|---|
| #1 (Three-class) | RF | 80.1 | **72.3** | 68.5 | 69.2 |
|  | CB | **80.3** | 72.2 | **70.0** | **70.5** |
|  | XB | 79.6 | 71.2 | 69.0 | 69.4 |
| #2 (Low vs Not Low) | RF | 91.7 | 90.5 | 91.9 | 91.1 |
|  | CB | **92.2** | **91.3** | **92.1** | **91.6** |
|  | XB | 91.8 | 90.7 | 91.9 | 91.3 |
| #3 (High vs Not High) | RF | 87.9 | 89.6 | **97.3** | 93.3 |
|  | CB | **88.9** | 91.1 | 96.6 | **93.8** |
|  | XB | 88.5 | **91.9** | 95.1 | 93.5 |
| **Acc**: Accuracy, **Pr**: Precision, **RF**: Random Forest, **CD**: CatBoost, **XB**: XGBoost | | | | | |

For FltSegment, SHAP values increase when the number of segments grows from very low counts up to ~100, but then decrease linearly, crossing into negative influence at around 200 segments. This result indicates that a moderate number of segments can improve structured, continuous descents, while an excessive number of short segments likely reflects degraded CDO performance.

Finally, for MDirection, SHAP values turn positive at approximately 1 and increase linearly up to ~2, after which they plateau. This suggests that maintaining the average directional changes below approximately 1° may reduce CDO performance. In contrast, moderate directional changes above 1°, up to about 2°, appear to support continuous descents and enhance CDO adherence, beyond which no further improvement is observed.

To further compare class separability across binary classification scenarios, the WD values were computed for the SHAP value distributions of each feature between the two classes. Table II summarizes these results, reporting the average WD across all features, the average WD for the five most discriminative features (i.e., those with the highest WD values), and the number of features with WD greater than 0.5.

As shown, the Low vs. Not Low scenario exhibits the largest WD values, indicating greater separation between classes in the feature importance space and, consequently, stronger discriminative potential. This finding is consistent with the ML classification results, where the Low vs. Not Low scenario achieved the highest accuracy across all classifiers.

Since the Low vs. Not Low classification demonstrated the best performance and class separability, and for simplification, we constructed and evaluated the FEXAI classifier only for this classification scenario, as described in the following section.

*C. FEXAI Classifier Results and Explainable Fuzzy Rules*

This section presents the classification results of the proposed FEXAI classifier along with the extracted fuzzy rules. To construct the FEXAI classifier, in the first step, three membership functions were designed for the top three most influential features, MDRate, FltSegments, and MDirection, based on domain knowledge and the SHAP value patterns identified in the previous step. Each membership function was defined with three fuzzy sets: Low, Medium, and High for MDRate; Few, Moderate, and Many for FltSegments; and Straight, Moderate, and Complex for MDirection as shown in Fig. 6. A winner-take-all approach was applied such that, for each feature value, the fuzzy set with the highest membership degree was selected. Using this approach and based on the intersections of the membership functions in Fig. 6, the selection criteria for each fuzzy set were then derived, as presented in Table III.

Using these membership functions, the FEXAI classifier was then constructed to classify CDOCAT using the top three features for the Low vs. Not Low scenario, as described in Section III. Table IV compares the classification results of the FEXAI classifier with those of the RF, CB, and XB classifiers, which were likewise trained and tested using only the top three features this time. While CB remained the best-performing classifier overall, FEXAI achieved competitive performance, outperforming RF and XB in all evaluation metrics except recall, where RF achieved the highest score.

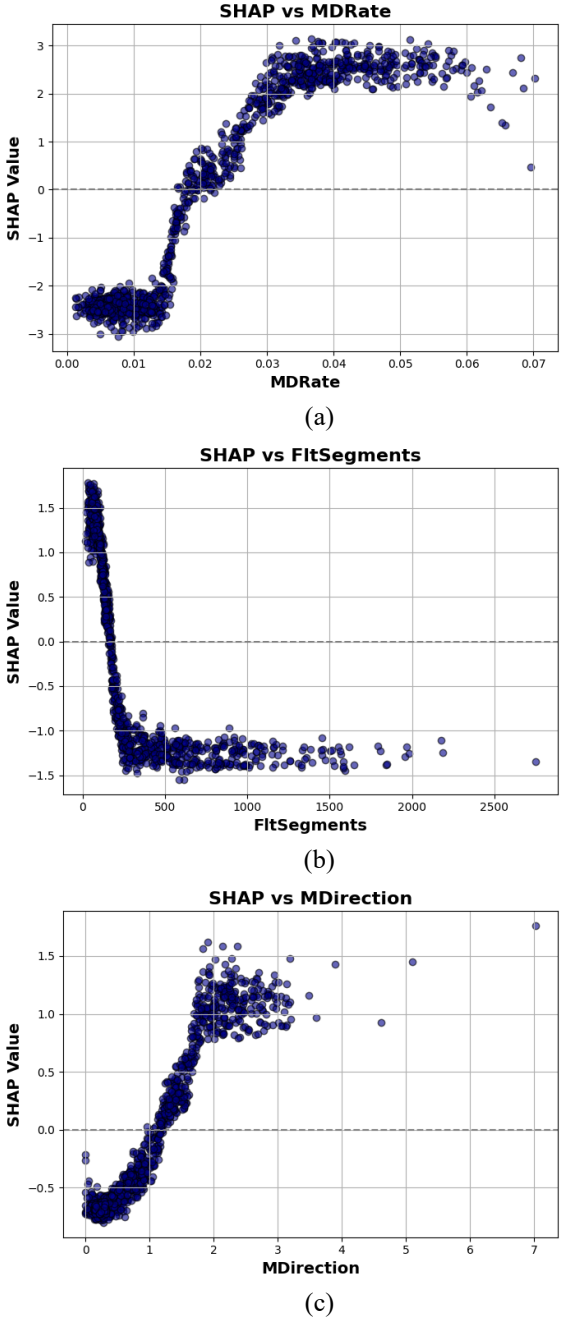

Fig. 5. Class-specific SHAP values for the top three features with highest SHAP scores, (a) MDrate, (b)FltSegments, and (c) MDirection.

TABLE II
MEAN WD VALUES BETWEEN SHAP DISTRIBUTIONS OF FEATURES ACROSS TWO CLASSES IN THE BINARY CLASSIFICATION SCENARIOS

| Scenario | Mean WD | Top 5 Mean WD | Nmber of features with WD>0.5 |
|---|---|---|---|
| **Low vs Not Low** | 0.21 | 1.28 | 3 |
| **High vs Not High** | 0.15 | 0.87 | 2 |

The extracted fuzzy rules are presented in Table V. These rules form an interpretable decision-making framework that is easily understandable by aviation operators, even without prior ML expertise. Notably, the fuzzy rules are also consistent with the SHAP-based insights from Fig. 5. For example, Rule 4 states that if MDRate, FltSegments, and MDirection are Medium, Few, and Moderate, respectively, then CDOCAT is classified as Not Low, which aligns with the SHAP class-specific patterns. Given the high interpretability of the FEXAI classifier, the slight reduction in performance compared to CB is a justified trade-off.

## V. Discussion

The results of this study demonstrated that combining SHAP-based feature attribution with fuzzy rule extraction through the proposed FEXAI method can produce both high classification performance and operational interpretability in assessing CDO adherence. To the best of our knowledge, this is the first study to classify CDO adherence and integrate fuzzy logic into CDO optimization, providing an explainable framework that is easily understandable to users with no prior knowledge of ML. Furthermore, the newly compiled dataset presented in this study is, to the best of our knowledge, the first comprehensive dataset specifically designed for CDO optimization.

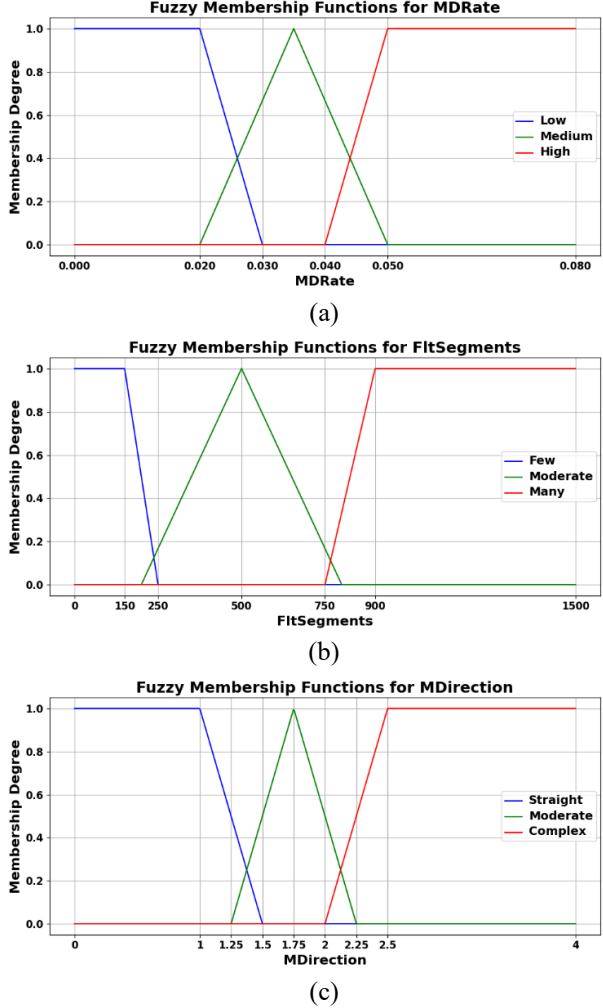

Fig. 6. Fuzzy membership functions for (a) MDRate, (b) FltSegments, and (c) MDirection.

TABLE III
PROJECTION OF DIFFERENT FEATURE VALUES TO CORRESPONDING FUZZY SETS

| Feature | Value range | Corresponding fuzzy set |
|---|---|---|
| MDR | MDR < 0.026 | Low |
|  | 0.026 < MDR < 0.044 | Medium |
|  | MDR > 0.044 | High |
| FS | FS < 238 | Few |
|  | 238 < FS < 767 | Moderate |
|  | FS > 767 | Many |
| MD | MD < 1.375 | Straight |
|  | 1.375 < MD < 2.125 | Moderate |
|  | MD > 2.125 | Complex |

**MDR**: MDRate, **FS**: FltSegments, **MD**: MDirection

TABLE IV
CLASSIFICATION PERFORMANCE OF ALL CLASSIFIERS USING ONLY THE TOP THREE FEATURES FOR LOW VS NOT LOW SCENARIO

| Classifier | Acc | Pr | Recall | F1 score |
|---|---|---|---|---|
| **FEXAI** | 90.5 | 89.5 | 89.9 | 89.7 |
| **RF** | 90.5 | 89.3 | **90.3** | 89.7 |
| **CB** | **90.8** | **90.0** | 90.2 | **89.9** |
| **XB** | 89.9 | 88.9 | 89.1 | 88.9 |

**Acc**: Accuracy, **Pr**: Precision, **RF**: Random Forest, **CB**: CatBoost, **XB**: XGBoost

TABLE V
EXTRACTED FUZZY RULED FROM THE FUZZY RULE-BASED CLASSIFIER

| # | Fuzzy rule |
|---|---|
| 1 | IF MDRate IS Low AND FltSegments IS Moderate AND MDirection IS Straight THEN CDOCAT IS **Low** |
| 2 | IF MDRate IS Low AND FltSegments IS Few AND MDirection IS Straight THEN CDOCAT IS Low |
| 3 | IF MDRate IS Low AND FltSegments IS Many AND MDirection IS Straight THEN CDOCAT IS Low |
| 4 | IF MDRate IS Medium AND FltSegments IS Few AND MDirection IS Moderate THEN CDOCAT IS Not Low |
| 5 | IF MDRate IS Medium AND FltSegments IS Few AND MDirection IS Complex THEN CDOCAT IS Not Low |
| 6 | IF MDRate IS Medium AND FltSegments IS Few AND MDirection IS Straight THEN CDOCAT IS Not Low |
| 7 | IF MDRate IS High AND FltSegments IS Few AND MDirection IS Moderate THEN CDOCAT IS Not Low |
| 8 | IF MDRate IS High AND FltSegments IS Few AND MDirection IS Straight THEN CDOCAT IS Not Low |
| 9 | IF MDRate IS High AND FltSegments IS Few AND MDirection IS Complex THEN CDOCAT IS Not Low |
| 10 | IF MDRate IS Low AND FltSegments IS Few AND MDirection IS Moderate THEN CDOCAT IS Not Low |
| 11 | IF MDRate IS Low AND FltSegments IS Few AND MDirection IS Complex THEN CDOCAT IS Not Low |
| 12 | IF MDRate IS Low AND FltSegments IS Moderate AND MDirection IS Moderate THEN CDOCAT IS Low |
| 13 | IF MDRate IS Low AND FltSegments IS Many AND MDirection IS Complex THEN CDOCAT IS Low |
| 14 | IF MDRate IS Low AND FltSegments IS Moderate AND MDirection IS Complex THEN CDOCAT IS Low |

CDO adherence was successfully classified using the compiled dataset, with all classifiers demonstrating high performance. The binary classification scenario distinguishing Low (CDO < 0.3) vs. Not Low (CDO ≥ 0.3) adherence yielded the highest performance across classifiers. This outcome is consistent with CDO behavior in real-world operations and previous studies [47]. Low CDO adherence typically reflects frequent level-offs, high-speed changes, and disrupted descent profiles caused by air traffic control interventions or inefficient arrival sequencing. These operational disruptions create more distinct feature patterns compared to flights with higher CDO adherence, where descent profiles tend to be smoother and more uniform [1]. As a result, the contrast between Low and Not Low categories is more pronounced, leading to improved model separability and classification performance.

The feature importance analysis revealed MDRate, FltSegments, and MDirection as the dominant predictors of CDO adherence across all scenarios. This finding is operationally significant. Higher median descent rates (MDRate) were generally associated with better CDO adherence up to an optimal range (~0.035), beyond which benefits plateaued or degraded. This aligns with ICAO guidance, which notes that excessively high descent rates may trigger ATC interventions or compromise passenger comfort, leading to operational inefficiencies [1].

FltSegments also emerged as a critical factor, with fewer segments generally indicating smoother, uninterrupted descents. This is consistent with ICAO guidelines [1] and previous studies [50], which state that segmentation of arrival routes often reflects ATC-imposed constraints or procedural interruptions, directly reducing CDO compliance. Our analysis suggests that segment counts below ~200 are indicative of higher adherence, aligning with operational goals of minimizing level-offs in the terminal area.

The influence of MDirection on CDO adherence in this study highlights the operational importance of moderate lateral path changes during descent. Our SHAP analysis indicates that average directional changes below approximately 1° are associated with lower CDO performance, whereas moderate changes between 1° and 2° support better adherence, after which performance plateaus. These findings are consistent with operational studies on point merge systems and arrival sequencing strategies [51], which have demonstrated that lateral maneuvering in terminal airspace can help maintain optimal descent profiles while accommodating traffic flow constraints.

The proposed FEXAI framework demonstrated clear interpretability advantages over existing models while maintaining high predictive performance. The extracted fuzzy rules offered a straightforward mapping for operational users to readily understand the decision logic. For practical deployment of FEXAI, two considerations are noteworthy. First, in this study, a five-fold cross-validation approach was employed to extract rules and evaluate the fuzzy classifier's accuracy. In an operational setting, however, the model could be trained on the full available dataset prior to deployment to maximize the robustness of the extracted rules. Second, the current implementation focused on the top three features identified through SHAP analysis for rule extraction. In practice, the number of features could be expanded, guided by domain expertise or targeted exploration of additional variables that, while not among the top-ranked features, may hold operational relevance in specific contexts. However, this should be balanced against the potential loss of interpretability that comes with incorporating more features into the rule base.

This study has several limitations. First, the dataset was restricted to flights on specific arrival routes into Doha Airport, which may limit the generalizability of the findings to other airports with different airspace structures, traffic densities, or ATC procedures. In particular, some features, such as MDirection, may vary considerably in other operational contexts. Second, the study did not address class imbalance in the dataset, which may have influenced model performance. Addressing this issue could further enhance the accuracy and robustness of all classifiers. Third, although the proposed method offered a clear advantage in interpretability and achieved the second-highest classification performance among the evaluated models, its slightly lower predictive accuracy compared to CatBoost should be acknowledged. The trade-off between transparency and accuracy should be carefully considered before operational deployment.

Future work could address these limitations by expanding the dataset to include a wider range of features and flights from different airports and operational environments. Correspondingly, the fuzzy membership functions may need adjustment to account for variations in feature distributions across new datasets. Further exploration of preprocessing and optimization techniques could also improve classification performance. Additionally, more advanced fuzzy inference approaches, such as Mamdani-type systems, could be investigated, provided that the high level of explainability demonstrated in this study is preserved.

REFERENCES

[1] ICAO, 'Continuous Descent Operations (CDO) Manual', International Civil Aviation Organization, Montréal, Doc 9931 AN/476, 2010.
[2] E. A. Alharbi, L. L. Abdel-Malek, R. J. Milne, and A. M. Wali, 'Analytical Model for Enhancing the Adoptability of Continuous Descent Approach at Airports', *Appl. Sci.*, vol. 12, no. 3, Art. no. 3, Jan. 2022.
[3] D. Toratani, N. K. Wickramasinghe, J. Westphal, and T. Feuerle, 'Feasibility study on applying continuous descent operations in congested airspace with speed control functionality: Fixed flight-path angle descent', *Aerosp. Sci. Technol.*, vol. 107, p. 106236, Dec. 2020.
[4] S. Kamo, J. Rosenow, H. Fricke, and M. Soler, 'Robust optimization integrating aircraft trajectory and sequence under weather forecast uncertainty', *Transp. Res. Part C Emerg. Technol.*, vol. 152, p. 104187, Jul. 2023.
[5] J.-P. B. Clarke *et al.*, 'Continuous Descent Approach: Design and Flight Test for Louisville International Airport', *J. Aircr.*, vol. 41, no. 5, pp. 1054–1066, Sep. 2004.
[6] J. Sun, J. Ellerbroek, and J. M. Hoekstra, 'Large-Scale Flight Phase Identification from ADS-B Data Using Machine Learning Methods', presented at the 7th International Conference on Research in Air Transportation, 2016.

[7] M. C. R. Murça, R. J. Hansman, L. Li, and P. Ren, 'Flight trajectory data analytics for characterization of air traffic flows: A comparative analysis of terminal area operations between New York, Hong Kong and Sao Paulo', *Transp. Res. Part C Emerg. Technol.*, vol. 97, pp. 324–347, Dec. 2018.

[8] S. Mondoloni and N. Rozen, 'Aircraft trajectory prediction and synchronization for air traffic management applications', *Prog. Aerosp. Sci.*, vol. 119, p. 100640, Nov. 2020.

[9] G. Jarry, D. Delahaye, F. Nicol, and E. Feron, 'Aircraft atypical approach detection using functional principal component analysis', *J. Air Transp. Manag.*, vol. 84, p. 101787, May 2020.

[10] X. Wu, H. Yang, H. Chen, Q. Hu, and H. Hu, 'Long-term 4D trajectory prediction using generative adversarial networks', *Transp. Res. Part C Emerg. Technol.*, vol. 136, p. 103554, Mar. 2022.

[11] M. Dai, 'A hybrid machine learning-based model for predicting flight delay through aviation big data', *Sci. Rep.*, vol. 14, no. 1, p. 4603, Feb. 2024.

[12] G. Demir, S. Moslem, and S. Duleba, 'Artificial Intelligence in Aviation Safety: Systematic Review and Biometric Analysis', *Int. J. Comput. Intell. Syst.*, vol. 17, no. 1, p. 279, Nov. 2024.

[13] S. Kamo, J. Rosenow, and H. Fricke, 'CDO Sensitivity Analysis for Robust Trajectory Planning under Uncertain Weather Prediction', in *2020 AIAA/IEEE 39th Digital Avionics Systems Conference (DASC)*, Oct. 2020, pp. 1–10.

[14] J.-P. B. Clarke et al., 'Continuous Descent Approach: Design and Flight Test for Louisville International Airport', *J. Aircr.*, vol. 41, no. 5, pp. 1054–1066, Sep. 2004.

[15] Y. Zhang and F. Chen, 'A Methodology for 4D Trajectory Optimization of Arrival Aircraft in Trajectory Based Operation', in *2018 13th World Congress on Intelligent Control and Automation (WCICA)*, Jul. 2018, pp. 1284–1289.

[16] Z. Xia et al., 'A systematic review on human-AI hybrid systems and human factors in air traffic management', *J. Eng. Des.*, vol. 0, no. 0, pp. 1–49.

[17] Z. Du, J. Wu, Y. Leng, and S. Wandelt, 'AI4ATM: A review on how Artificial Intelligence paves the way towards autonomous Air Traffic Management', *J. Air Transp. Res. Soc.*, vol. 5, p. 100077, Dec. 2025.

[18] S. M. Lundberg and S.-I. Lee, 'A unified approach to interpreting model predictions', in *Proceedings of the 31st International Conference on Neural Information Processing Systems*, in NIPS'17. Red Hook, NY, USA: Curran Associates Inc., Dec. 2017, pp. 4768–4777.

[19] A. Noroozi, S. M. Esha, and M. Ghari, 'Predictors of Childhood Vaccination Uptake in England: An Explainable Machine Learning Analysis of Longitudinal Regional Data (2021-2024)', Apr. 18, 2025, *arXiv*: arXiv:2504.13755.

[20] Y. Zhou, S. Booth, M. T. Ribeiro, and J. Shah, 'Do Feature Attribution Methods Correctly Attribute Features?', *Proc. AAAI Conf. Artif. Intell.*, vol. 36, no. 9, Art. no. 9, Jun. 2022.

[21] B. Kim et al., 'Interpretability Beyond Feature Attribution: Quantitative Testing with Concept Activation Vectors (TCAV)', presented at the International Conference on Machine Learning, Nov. 2017.

[22] F. Aghaeipoor, M. Sabokrou, and A. Fernández, 'Fuzzy Rule-Based Explainer Systems for Deep Neural Networks: From Local Explainability to Global Understanding', *IEEE Trans. Fuzzy Syst.*, vol. 31, no. 9, pp. 3069–3080, Sep. 2023.

[23] A. Noroozi, R. P. Hasanzadeh, and M. Ravan, 'A fuzzy alignment approach for identification of arbitrary crack shape profiles in metallic structures using ACFM technique', in *20th Iranian Conference on Electrical Engineering (ICEE2012)*, May 2012, pp. 894–899.

[24] A. Noroozi, R. P. R. Hasanzadeh, and M. Ravan, 'A Fuzzy Learning Approach for Identification of Arbitrary Crack Profiles Using ACFM Technique', *IEEE Trans. Magn.*, vol. 49, no. 9, pp. 5016–5027, Sep. 2013.

[25] T. Fasoro, 'Trajectory Specification to Support High-Throughput Continuous Descent Approaches', Thesis, Massachusetts Institute of Technology, 2022.

[26] J. M. Canino Rodriguez, L. Gomez Deniz, J. Garcia Herrero, J. Besada Portas, and J. R. Casar Corredera, 'A 4D trajectory negotiation protocol for Arrival and Approach sequencing', in *2008 Integrated Communications, Navigation and Surveillance Conference*, May 2008, pp. 1–12.

[27] R. Sáez, X. Prats, T. Polishchuk, V. Polishchuk, and C. Schmidt, 'Automation for Separation with Continuous Descent Operations: Dynamic Aircraft Arrival Routes', *J. Air Transp.*, vol. 28, no. 4, pp. 144–154, Oct. 2020.

[28] A. S. Rohani, T. G. Puranik, and K. M. Kalyanam, 'Machine Learning Approach for Aircraft Performance Model Parameter Estimation for Trajectory Prediction Applications', in *2023 IEEE/AIAA 42nd Digital Avionics Systems Conference (DASC)*, Oct. 2023, pp. 1–9.

[29] N. Takeichi, 'Adaptive prediction of flight time uncertainty for ground-based 4D trajectory management', *Transp. Res. Part C Emerg. Technol.*, vol. 95, pp. 335–345, Oct. 2018.

[30] C. Ma, D. Mookherjee, G. Mesquida-Masana, et al., 'A Top of Descent Prediction Model for Interaction-Free Continuous Descent Operations', presented at the the SESAR Innovation Days (SIDs), 2024.

[31] S. Baumann and U. Klingauf, 'Modeling of aircraft fuel consumption using machine learning algorithms', *CEAS Aeronaut. J.*, vol. 11, no. 1, pp. 277–287, Jan. 2020.

[32] W. A. Khan, H.-L. Ma, X. Ouyang, and D. Y. Mo, 'Prediction of aircraft trajectory and the associated fuel consumption using covariance bidirectional extreme learning machines', *Transp. Res. Part E Logist. Transp. Rev.*, vol. 145, p. 102189, Jan. 2021.

[33] W. A. Khan, H.-L. Ma, X. Ouyang, and D. Y. Mo, 'Prediction of aircraft trajectory and the associated fuel consumption using covariance bidirectional extreme learning machines', *Transp. Res. Part E Logist. Transp. Rev.*, vol. 145, p. 102189, Jan. 2021.

[34] Y. Zhao, Z. Wang, X. Wang, Y. Song, and Y. Han, 'Data driven fuel consumption prediction model for green aviation using radial basis function neural network', *Sci. Rep.*, vol. 15, no. 1, p. 26275, Jul. 2025.

[35] C. S. Hernandez, S. Ayo, and D. Panagiotakopoulos, 'An Explainable Artificial Intelligence (xAI) Framework for Improving Trust in Automated ATM Tools', in *2021 IEEE/AIAA 40th Digital Avionics Systems Conference (DASC)*, Oct. 2021, pp. 1–10.

[36] A. Nanyonga, H. Wasswa, K. Joiner, U. Turhan, and G. Wild, 'Explainable Supervised Learning Models for Aviation Predictions in Australia', *Aerospace*, vol. 12, no. 3, Art. no. 3, Mar. 2025.

[37] A. Khattak, J. Zhang, P. Chan, and F. Chen, 'SPE-SHAP: Self-paced ensemble with Shapley additive explanation for the analysis of aviation turbulence triggered by wind shear events', *Expert Syst. Appl.*, vol. 254, p. 124399, Nov. 2024.

[38] M. U. Ahmed et al., 'Role of Multi-modal Machine Learning, Explainable AI and Human-AI Teaming in Trusted Intelligent Systems for Remote Digital Towers', in *Proceedings of the 2024 7th Artificial Intelligence and Cloud Computing Conference*, in AICCC '24. New York, NY, USA: Association for Computing Machinery, Jul. 2025, pp. 26–35.

[39] M. S. Raboaca, C. Dumitrescu, and I. Manta, 'Aircraft Trajectory Tracking Using Radar Equipment with Fuzzy Logic Algorithm', *Mathematics*, vol. 8, no. 2, Art. no. 2, Feb. 2020.

[40] I. Dagal, W. F. Mbasso, H. Ambe, B. Erol, and P. Jangir, 'Adaptive Fuzzy Logic Control Framework for Aircraft Landing Gear Automation: Optimized Design, Real-Time Response, and Enhanced Safety', *Int. J. Aeronaut. Space Sci.*, Mar. 2025.

[41] B. Burchett, 'Fuzzy Logic Trajectory Design and Guidance for Terminal Area Energy Management'. Apr. 01, 2003.

[42] A. Sathyan, N. Ernest, L. Lavigne, F. Cazaurang, M. Kumar, and K. Cohen, 'A Genetic Fuzzy Logic Based Approach to Solving the Aircraft Conflict Resolution Problem', in *AIAA Information Systems-AIAA Infotech @ Aerospace*, in AIAA SciTech Forum., American Institute of Aeronautics and Astronautics, 2017.

[43] A. Noroozi, M. S. Hasan, M. Ravan, E. Norouzi, and Y.-Y. Law, 'An efficient machine learning approach for extracting eSports players' distinguishing features and classifying their skill levels using symbolic transfer entropy and consensus nested cross-validation', *Int. J. Data Sci. Anal.*, Mar. 2024.

[44] M. Margarette Sanchez et al., 'A Machine Learning Algorithm to Discriminating Between Bipolar and Major Depressive Disorders Based on Resting EEG Data', in *2022 44th Annual International Conference of the IEEE Engineering in Medicine & Biology Society (EMBC)*, Jul. 2022, pp. 2635–2638.

[45] M. Ravan et al., 'Discriminating between bipolar and major depressive disorder using a machine learning approach and resting-state EEG data', *Clin. Neurophysiol.*, vol. 146, pp. 30–39, Feb. 2023.

[46] H. Hardell, A. Lemetti, T. Polishchuk, L. Smetanová, and K. Zeghal, 'Towards a Comprehensive Characterization of the Arrival Operations in the Terminal Area', presented at the 11th SESAR Innovation Days (SIDs), Virtual, December 7-9, 2021, 2021.

[47] T. Huynh, A. Thomas, and J. Bronsvoort, 'ANSP Measures of Flight Descent Performance An Evaluation of CDO and Managed Descent', presented at the 11th SESAR Innovation Days, 2021.

[48] Y. Rubner, C. Tomasi, and L. J. Guibas, 'The Earth Mover's Distance as a Metric for Image Retrieval', *Int. J. Comput. Vis.*, vol. 40, no. 2, pp. 99–121, Nov. 2000.

[49] G. Casalino, G. Castellano, C. Castiello, and C. Mencar, 'Effect of fuzziness in fuzzy rule-based classifiers defined by strong fuzzy partitions and winner-takes-all inference', *Soft Comput.*, vol. 26, no. 14, pp. 6519–6527, Jul. 2022.

[50] T. J. Callantine, M. Kupfer, L. H. Martin, and T. Prevot, 'Simulations of Continuous Descent Operations with Arrival-management Automation and Mixed Flight-deck Interval Management Equipage', presented at the USA/Europe Air Traffic Management R&D Seminar (ATC2013), Chicago, IL, Jun. 2013.

[51] D. Luo, 'Evaluation of Airspace Operation Safety Level based on PMS', *E3S Web Conf.*, vol. 512, p. 03019, 2024.